\documentclass{article}

\usepackage{arxiv}

\usepackage[utf8]{inputenc} 
\usepackage[T1]{fontenc}    
\usepackage{hyperref}       
\usepackage{url}            
\usepackage{booktabs}       
\usepackage{amsfonts}       
\usepackage{nicefrac}       
\usepackage{microtype}      
\usepackage{lipsum}
\usepackage{graphicx}
\usepackage{subfig}
\usepackage{amsmath}
\usepackage{algorithm}
\usepackage{algpseudocode}

\title{Supervised Learning Approaches for Static and Temporal Force Prediction in a Biohybrid Muscle Ring Actuator}

\author{
    Michail-Antisthenis Tsompanas  \\
    Unconventional Computing Laboratory \& \\
    School of Computing \& Creative Technologies,\\
    College of Arts, Technology and Environment, \\
    University of the West of England,\\ 
    Bristol, BS16 1QY, United Kingdom \\
    \texttt{antisthenis.tsompanas@uwe.ac.uk}
\And
    Marco Perez Hernandez, Faisal Abdul-Fattah, Karim Elhakim, Mostafa Ibrahim \\
    School of Computing \& Creative Technologies,\\
    College of Arts, Technology and Environment, \\
    University of the West of England,\\ 
    Bristol, BS16 1QY, United Kingdom 
\And
    Judith Fuentes, Florencia Lezcano \\
    Institute for Bioengineering of Catalonia (IBEC), \\ 
    The Barcelona Institute for Science and Technology (BIST),\\ Barcelona, Spain
\And
    Riccardo Collu, Massimo Barbaro, Stefano Lai \\
    Department of Electrical and Electronic Engineering, \\
    University of Cagliari,\\
    Cagliari, Italy
\And
      Samuel S{\'a}nchez\\
       Instituciò Catalana de Recerca i Estudis Avancats (ICREA) and \\ 
      Institute for Bioengineering of Catalonia (IBEC), \\ 
    The Barcelona Institute for Science and Technology (BIST),\\ Barcelona, Spain
\And
       Andrew Adamatzky \\
       Unconventional Computing Laboratory,\\
    College of Arts, Technology and Environment, \\
    University of the West of England,\\ 
    Bristol, BS16 1QY, United Kingdom \\
}

\begin{document}
\maketitle

\begin{abstract}
Skeletal muscle-based biohybrid actuators have proved to be a promising component in soft robotics, offering efficient movement. However, their intrinsic biological variability and nonlinearity pose significant challenges for controllability and predictability. To address these issues, this study investigates the application of supervised learning, a form of machine learning, to model and predict the behavior of biohybrid machines (BHMs), focusing on a muscle ring anchored on flexible polymer pillars. First, static prediction models (i.e., random forest and neural network regressors) are trained to estimate the maximum exerted force achieved from input variables such as muscle sample, electrical stimulation parameters, and baseline exerted force. Second, a dynamic modeling framework, based on Long Short-Term Memory networks, is developed to replicate the time series of exerted forces observed. Both modeling approaches demonstrate high predictive accuracy. The best performance of the static models is characterized by a mean $R^2$ of 0.9346, whereas the dynamic model achieves a mean $R^2$ of 0.9944 over five training instances. The static models can enable optimization of muscle actuator performance for targeted applications and required force outcomes, while the dynamic model provides a foundation for developing robustly adaptive control strategies in future biohybrid robotic systems.
\end{abstract}


\section{Introduction}\label{sec1}
Soft robotics is a rapidly expanding field of robotics that aims at creating lifelike machines capable of adaptable and safe interactions with complex environments. Several options are available for mobilizing soft robotics, although a promising approach is Biohybrid Machines (BHMs). BHMs integrate living biological components such as muscle cells or engineered tissues with synthetic, flexible materials \cite{ricotti2012bio,ricotti2017biohybrid}. By merging the efficiency and functional intelligence of living matter with the reliability and robustness of engineered substrates, BHMs can achieve movement and adaptability that traditional robotic actuators, like pneumatic, hydraulic, or dielectric, struggle to replicate \cite{bartolucci2025monolithic}. These biohybrid systems have shown particular promise in applications requiring compliance and bio-compatibility, including soft robotic manipulators, minimally invasive medical instruments \cite{cianchetti2018biomedical}, and adaptive biomedical devices designed for targeted drug deposition \cite{joyee2020additive}.

Engineered skeletal muscle tissues are integrated to BHMs, since they are capable of controllable and reversible contractions, offer high energy efficiency, intrinsic self-healing capabilities, and consistent performance across scales \cite{webster2022biohybrid,appiah2019living}. Such characteristics make them particularly suitable for generating biomimetic motions such as grasping \cite{kabumoto2013voluntary}, walking \cite{kinjo2024biohybrid} or swimming \cite{guix2021biohybrid}. Their miniaturization potential also opens avenues for precise manipulation within delicate environments, such as the vascular system or other confined anatomical spaces \cite{bartolucci2025monolithic}. By integrating muscle-actuated flexures with soft polymeric structures and organic electronics, researchers are now approaching levels of performance previously unattainable in conventional soft actuators, enabling unprecedented motion control and adaptability \cite{bartolucci2025monolithic,higueras2021artificial}.

However, despite these advances, the reliable actuation and control of BHMs remain major challenges \cite{li2025advancing}. Biological variability, tissue fatigue, and the intrinsic nonlinear behavior of pliable materials make it difficult to predict responses to external stimuli \cite{collu2025development}. Moreover, the stimulation methods that often rely on electrical, magnetic, optical, or chemical cues introduce further complexity \cite{carlsen2014bio}. Conventional electrical stimulation techniques based on voltage control frequently fail to ensure precise charge delivery, as their effectiveness depends strongly on electrode impedance, medium composition, and geometric factors \cite{merrill2005electrical}. This lack of predictability complicates the replication of results and limits the scalability of biohybrid designs. In addition, overstimulation can lead to cellular stress and tissue degradation, reducing actuator lifespan and performance stability. As a result, conventional control models, which assume linearity and static system parameters, are inadequate for managing the dynamic, complex, and time-varying behavior of living actuators.

In recent years, {Machine Learning} (ML) has emerged as a transformative tool to address these limitations \cite{chi2024perspective,wang2022control}. ML-based models are capable of learning and representing complex, nonlinear relationships between stimulation inputs and mechanical outputs, enabling accurate prediction and real-time control of biohybrid systems \cite{ramdya2023neuromechanics}. Beyond black-box prediction, advances in explainable ML will allow researchers to extract interpretable features and understand the underlying dynamics governing engineered muscle actuator behavior. Through such approaches, it becomes possible to not only optimize stimulation protocols, by adjusting amplitude, frequency, or waveform characteristics, but also to design adaptive feedback control strategies through sensing \cite{mahmood2024integration,filippi2024sensor,lai2024real} that will be able to respond autonomously to variations in biological performance. This capability is crucial for advancing BHMs from proof-of-concept demonstrations toward reliable, functional robotic devices capable of sustained operation in biomedical or environmental settings \cite{sarker2025review}.

In this work, we propose two ML-based approaches to explain and predict the behavior of muscle-actuated mechanisms. These methods are trained on experimental data acquired from a custom-designed, programmable stimulation system applied on a simple two-post set-up where a muscle ring is anchored between the pillars. The dataset represents the interplay between electrical excitation and muscle exerted force, which is calculated based on the mechanical deformation of the pillars \cite{collu2025development}. To address the key challenges of variability and nonlinearity in biohybrid actuation, we explore two complementary modeling strategies. The first approach focuses on static prediction, employing at first random forest (RF) and then feedforward neural network (NN) models to estimate the maximum possible exerted force on the pillars based on the muscle sample, electrical stimulation parameters, and baseline behavior before the application of stimuli. The second approach introduces a dynamic modeling framework, utilizing a recurrent NN, specifically a Long Short-Term Memory (LSTM) architecture, to construct {a data-driven predictive model} of the biohybrid mechanism capable of reproducing its full time series exerted force throughout experiments with electrical stimulation. Namely, this scheme is studied as a univariate time series forecasting problem. Both approaches demonstrate high predictive accuracy, offering valuable insights into the underlying actuation dynamics and establishing a foundation for the development of adaptive control strategies in future ML-driven biohybrid robotic systems.

\section{Results and Discussion}\label{sec4}

{The dataset was constructed using 15 biological replicates of muscle ring actuators (or samples) anchored in a two-post setup (as illustrated in Fig. \ref{fig_ric}C). In total, 161 experiments were conducted, where individual samples were stimulated under specific electrical conditions (frequency, pulse width, and waveform). Some experimental conditions were repeated to verify consistency in the measured force output and to mitigate potential sample defects. The resulting dataset therefore reflects both variability across different actuator samples and across stimulation conditions (for further details of the inputs and outputs please refer to Section \ref{Static_methods}). The experiments can be categorized according to frequency and pulse width, as depicted in Tables \ref{tab1} and \ref{tab2}. An overview of all stimulation conditions is provided in Table \ref{stim_summary}.}

\begin{table}[!h]%
\centering %
\caption{Stimulation Frequency distribution on the dataset.\label{tab1}}%
\begin{tabular}{cc}
\toprule
\textbf{Stimulation Frequency} & \textbf{Amount of experiments}  \\
\midrule
50 & 69  \\
30  & 48  \\
20  & 30  \\
35  &  12  \\
70  &   1  \\
25  &   1  \\
\bottomrule
\end{tabular}
\end{table}

\begin{table}[!h]%
\centering %
\caption{Stimulation Pulse Width distribution on the dataset.\label{tab2}}%
\begin{tabular}{cc}
\toprule
\textbf{Stimulation Pulse Width} & \textbf{Amount of experiments}  \\
\midrule
15 &  53 \\
5  &  41  \\
20 &  17 \\
10 &  15 \\
8  &  13  \\
40 &  10 \\
30 &  9  \\
28 &  3  \\
\bottomrule
\end{tabular}
\end{table}

\begin{figure}
\centerline{\includegraphics[width=0.5\linewidth]{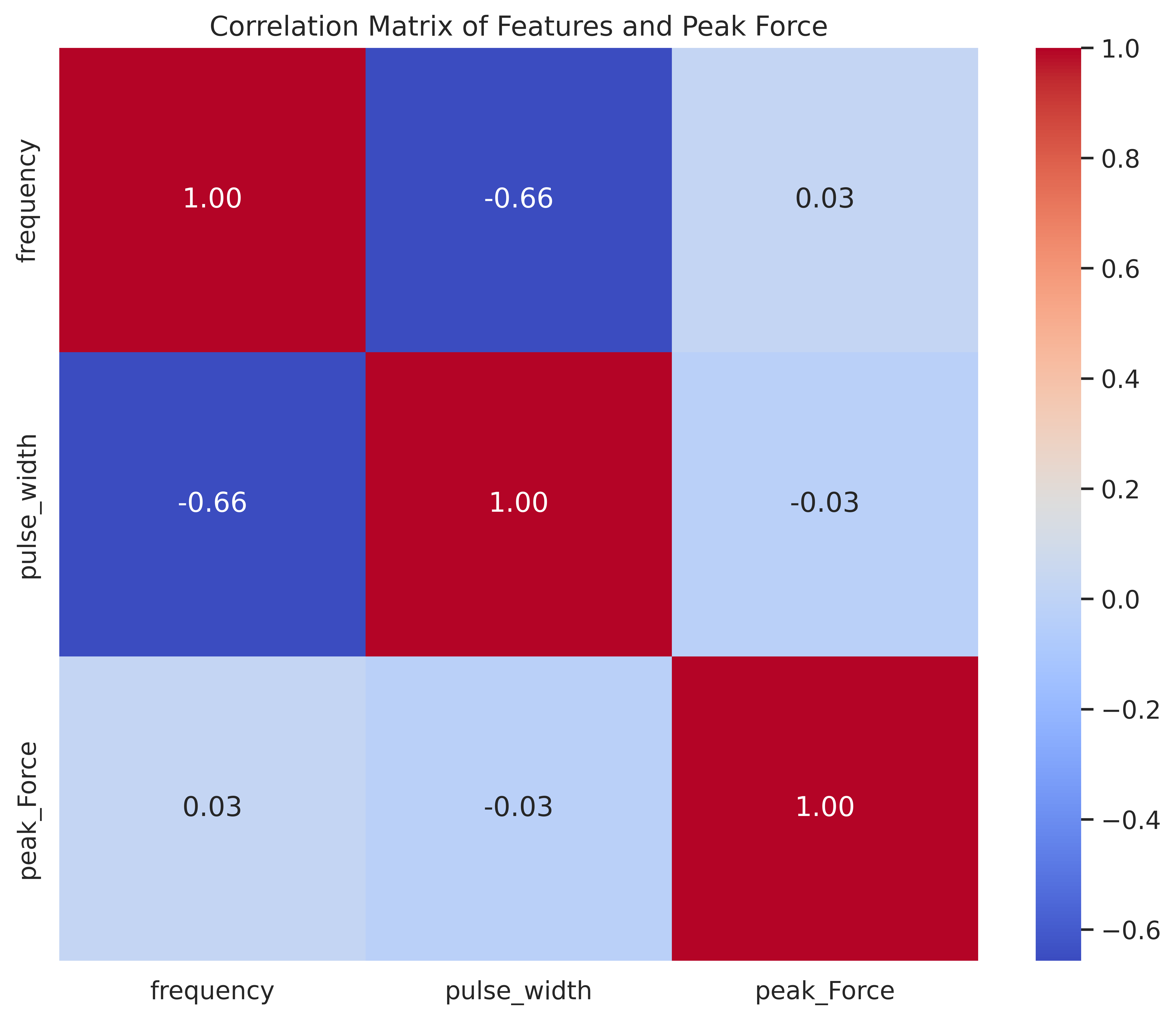}}
\caption{Correlation matrix of numerical features and output - peak\_Force.\label{in_cor}}
\end{figure}

{For the static prediction models, the dataset of 161 data points (with four inputs -sample number ID, waveform type, frequency and pulse width- and one output -the maximum force achieved per experiment) was utilized. To better illustrate the relations between inputs and output, the Pearson correlation between numerical inputs (frequency and pulse width) and the output are provided in Fig. \ref{in_cor}. Moreover, for categorical inputs the correlation ratio ($\eta$) between sample number ID and output is $0.97$, whereas between waveform type and output is $0.88$. This exploratory analysis of the dataset revealed that the biological sample number ID is the dominant factor influencing the generated force. This highlights the substantial intrinsic variability among muscle rings and indicates that stimulation parameters alone cannot adequately explain the force output. Consequently, the sample identifier was included as an input feature to account for this biological variability. While this improves predictive performance, it also limits direct generalization to unseen actuators, as the model effectively learns sample-specific characteristics. In practical applications, this limitation could be addressed through an initial calibration procedure for new biological samples or by replacing the sample identifier with measurable actuator-specific descriptors.}

The dataset was divided into 80\% for training and 20\% for testing. Specifically, the training set consisted of 128 data points, while the testing set contained 33 data points. 

\begin{figure*}
\centerline{\includegraphics[width=0.8\linewidth]{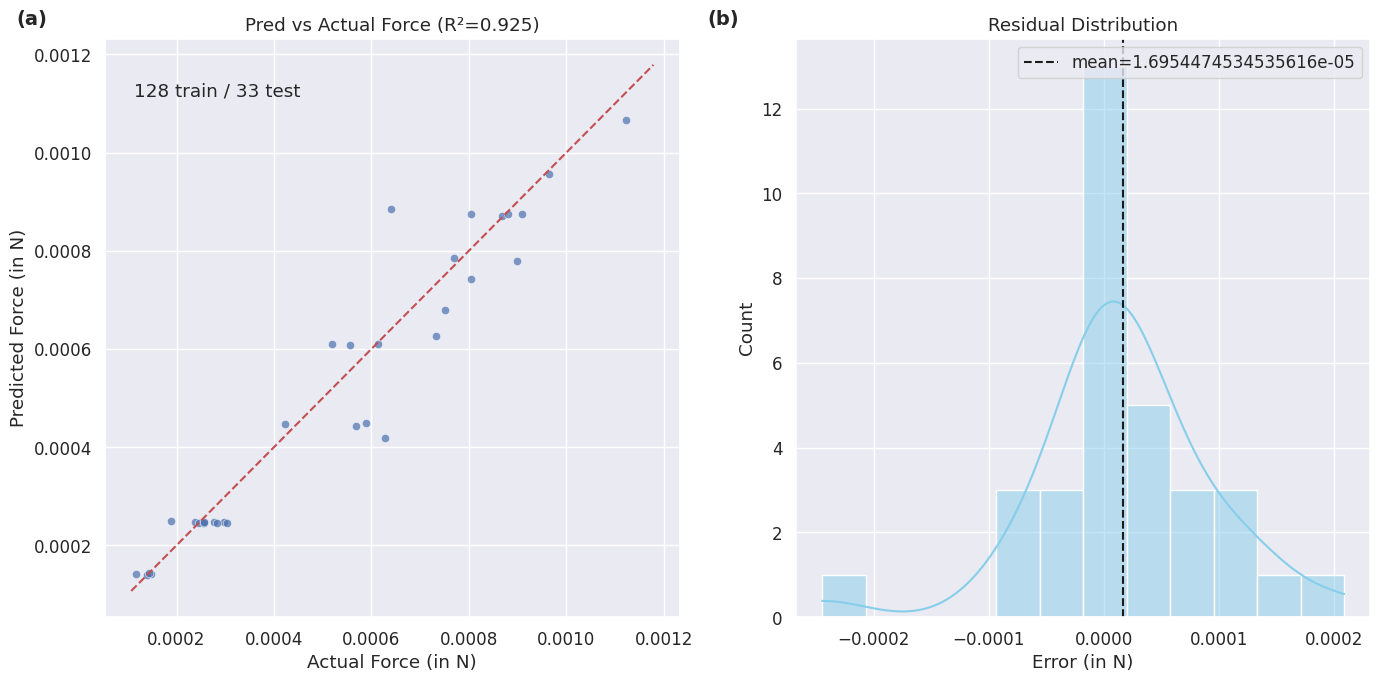}}
\caption{RFR model accuracy and residual analysis for predicted force, (A) scatter plot of actual versus predicted values and (B) residual distribution.\label{fig1}}
\end{figure*}

The performance of the Random Forest Regressor (RFR) was optimized through hyperparameter tuning, via randomized search cross validation. {The RFR model was trained with five different random seeds and achieved its best performance with the following hyperparameters: number of estimators equal to 200, maximum depth of individual trees equal to 10, minimum samples split equal to 2 and minimum samples leaf equal to 1. The optimized model fitted the dataset with a root mean square error (RMSE) of $8.249 \times 10^{-5} \pm 3.895 \times 10^{-6}$  and an $R^2$ score of $0.9183 \pm 0.0078$. The scatter plot of actual versus predicted force by one of the trained RFR models and the corresponding residual analysis are illustrated in Fig. \ref{fig1}.} The RFR model demonstrated adequate but limited predictive accuracy, which can be attributed to the limited dataset size and the inherent biological variability among muscle rings, even though identical fabrication and maturing procedures were followed.

\begin{figure*}
\centerline{\includegraphics[width=0.8\linewidth]{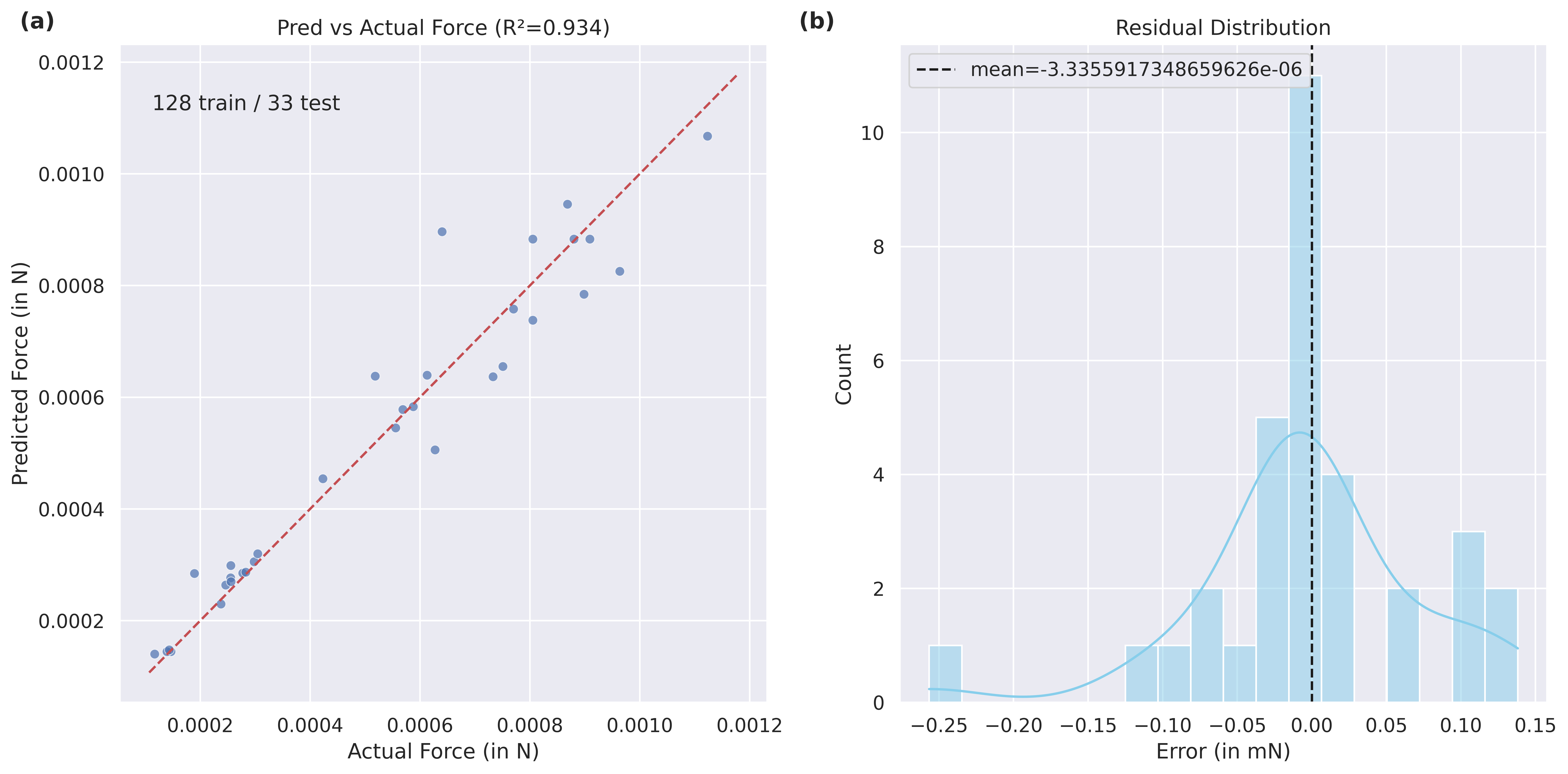}}
\caption{NN model accuracy and residual analysis for predicted force, (A) scatter plot of actual versus predicted values and (B) residual distribution.\label{fig2}}
\end{figure*}

Since, NNs are better suited to capture complex and nonlinear relationships between inputs and outputs, a NN with an architecture of four hidden layers with 68, 50, 30, and 10 neurons respectively, was implemented. It is noteworthy here that the same inputs and output are considered for the NN as with the RFR. {The NN was trained with five different random seeds and achieved an RMSE of $7.642 \times 10^{-5} \pm 3.202 \times 10^{-6}$  and an $R^2$ score of $0.9299 \pm 0.0058$. The scatter plot of actual versus predicted force by one of the NN models and the residual analysis are illustrated in Fig. \ref{fig2}.} The slightly improved accuracy of the NN relative to the RFR is consistent with the expected cross-feature dependencies present in the dataset.

\begin{figure*}
\centerline{\includegraphics[width=0.8\linewidth]{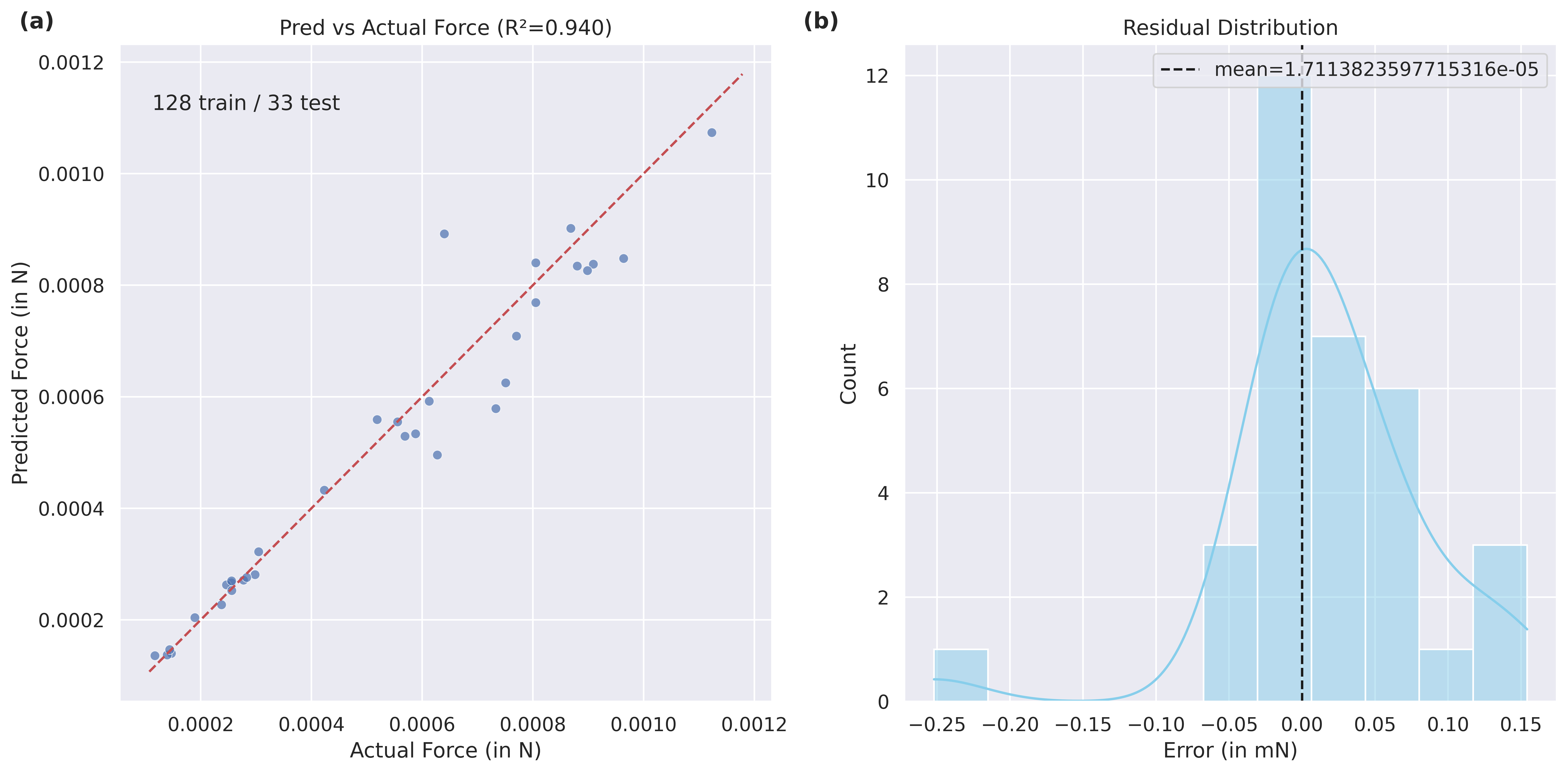}}
\caption{Updated NN model accuracy and residual analysis for predicted force, (A) scatter plot of actual versus predicted values and (B) residual distribution.\label{fig3}}
\end{figure*}

To further enhance the NN model performance and mitigate the effect of high biological variability, an additional input feature was introduced. Namely, the average of the first 10 measurements (equivalent to the first 0.4 seconds in total) recorded for each experiment. During this initial period, no electrical stimulation was applied. Moreover, the architecture of the network in terms of hidden neuron layers and hyperparamenters, like optimizer type and training epochs, were maintained the same as the previous application of NN. {With this additional input, the updated NN was trained with five different random seeds and achieved an RMSE of $7.375 \times 10^{-5} \pm 4.151 \times 10^{-6}$  and an $R^2$ score of $0.9346 \pm 0.0075$. The scatter plot of actual versus predicted forces by one of the updated NN models and the corresponding residual analysis of the dataset are shown in Fig. \ref{fig3}.} Incorporating the baseline exerted force prior to stimulation further improved model accuracy, as expected. It is noteworthy here that the NN with additional input feature of the unstimulated force reduced RMSE by 3.5\% relative to the initial NN model and by 10.6\% relative to the RFR model.

The residual distribution of the RFR model is symmetrically distributed around zero, with a mean of $1.69\times 10^{-5} N$ and only a few outliers (i.e., around $\pm 0.0002 N$). These outliers correspond to samples with difficult to predict force values. Similar results are observed for the initial NN model (a mean of $-3.33\times 10^{-6} N$). The updated NN model demonstrated a similar residual spread (i.e., mean equal to $1.71\times 10^{-5} N$), however, with the outlier residuals closer to zero. This suggests that the inclusion of baseline force as an additional input effectively reduced systematic variability and improved the model’s ability to capture nonlinear dependencies in the dataset. Moreover, for all static models the residual distribution does not reveal any trend in the errors of predictions.

\begin{figure*}
\centerline{\includegraphics[width=0.8\linewidth]{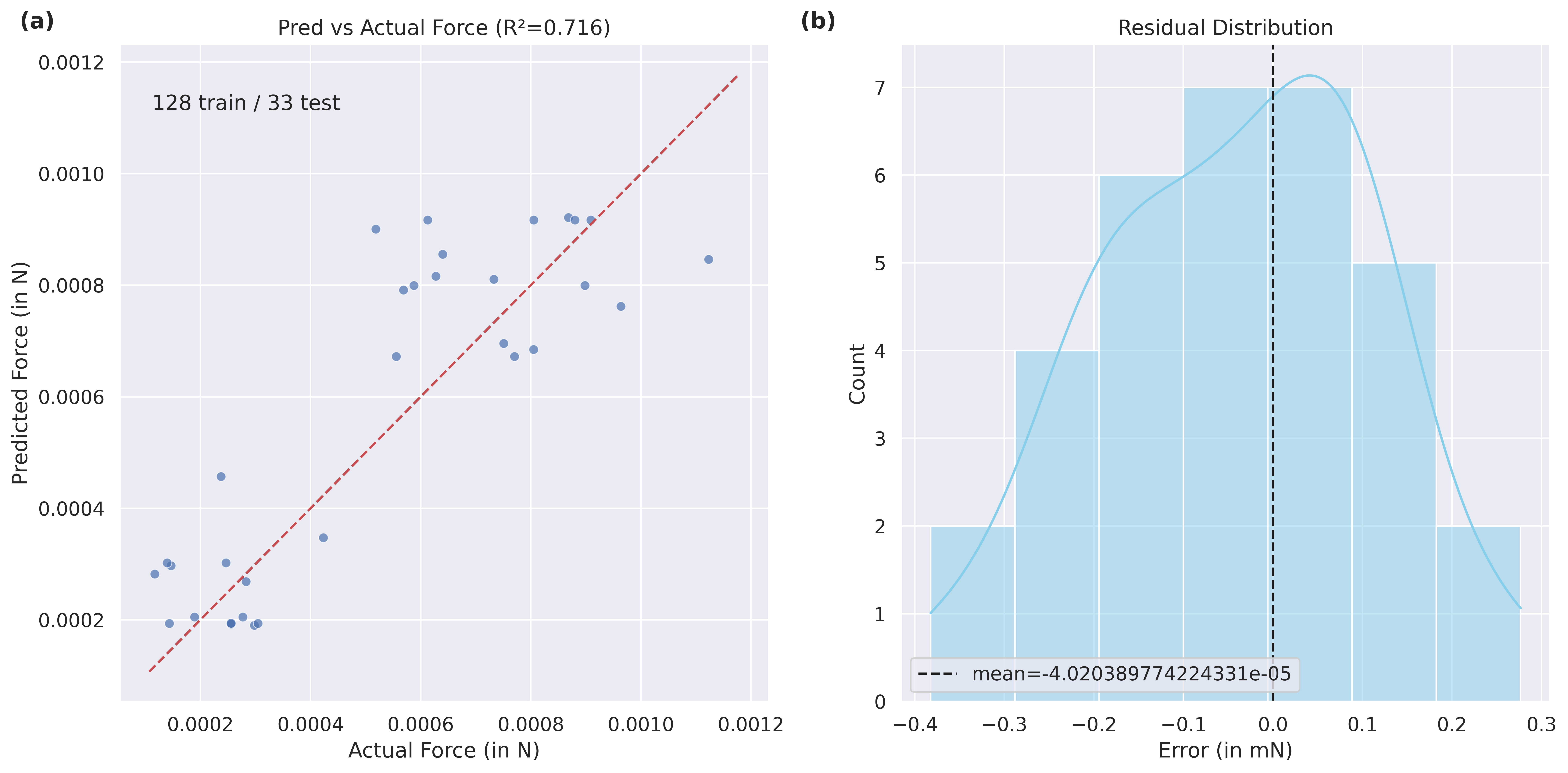}}
\caption{Updated NN model accuracy and residual analysis for predicted force without the sample number ID input, (A) scatter plot of actual versus predicted values and (B) residual distribution.\label{fig2b}}
\end{figure*}

\begin{figure*}
\centerline{\includegraphics[width=0.8\linewidth]{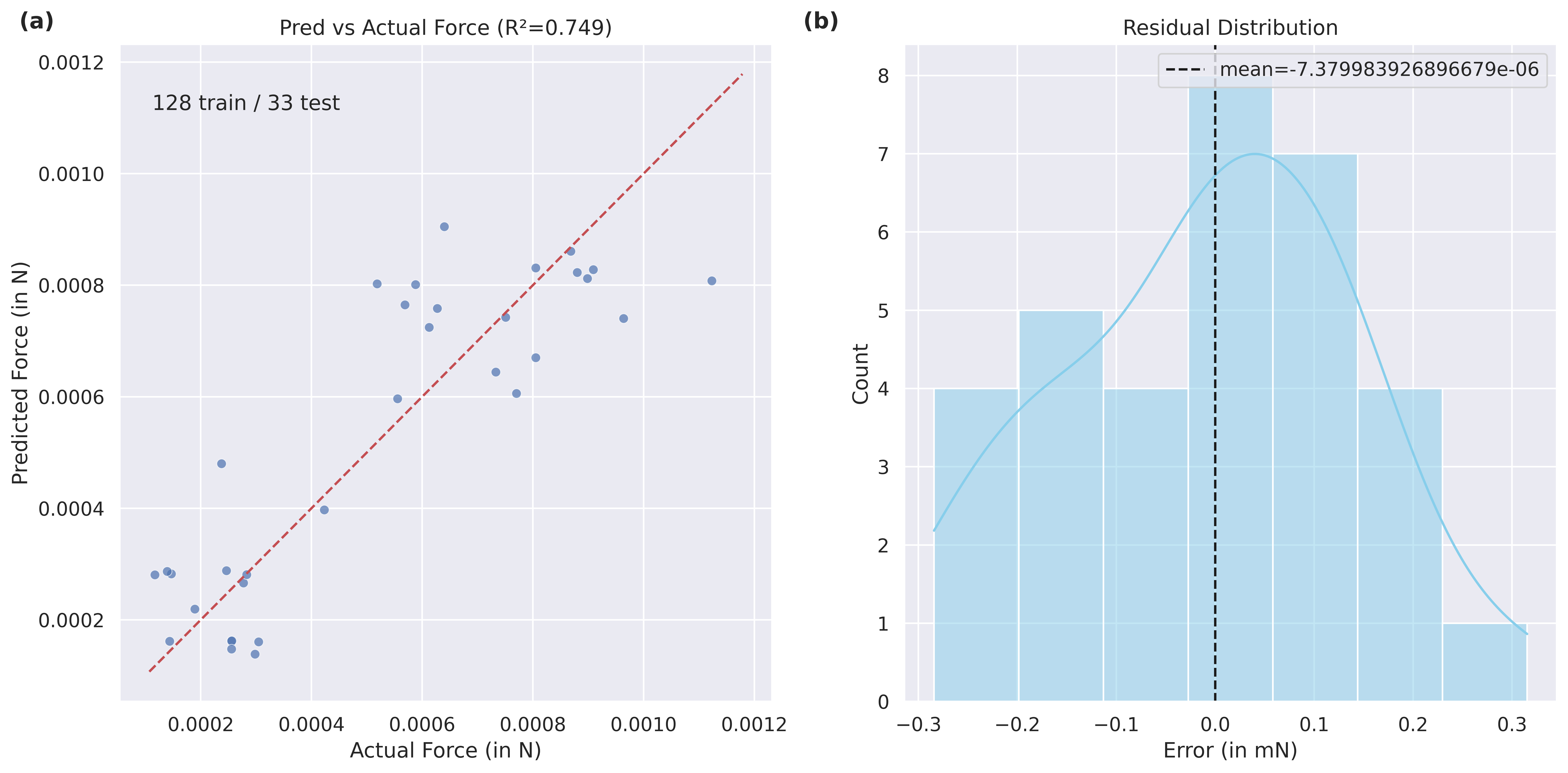}}
\caption{Updated NN model accuracy and residual analysis for predicted force without the sample number ID input, (A) scatter plot of actual versus predicted values and (B) residual distribution.\label{fig3b}}
\end{figure*}

{To investigate the influence of the biological sample identifier on the predictive performance, the two best performing static models (NN and updated NN) were additionally trained without using the muscle ring sample identifier as an input feature. The NN was trained again with five different random seeds and achieved an RMSE of $0.00015 \pm 3.43 \times 10^{-6}$  and an $R^2$ score of $0.7204 \pm 0.0126$. The scatter plot of actual versus predicted force by one of the NN models and the residual analysis are illustrated in Fig. \ref{fig2b}. Whereas, the updated NN was trained again with five different random seeds and achieved an RMSE of $0.00014 \pm 4.58 \times 10^{-6}$  and an $R^2$ score of $0.7409 \pm 0.0163$. The scatter plot of actual versus predicted force by one of the NN models and the residual analysis are illustrated in Fig. \ref{fig3b}. As expected, both models exhibited a noticeable decrease in predictive accuracy, confirming that biological variability between muscle rings is a dominant factor affecting the generated force. This result indicates that stimulation parameters alone do not fully determine the actuator response. Although removing the sample identifier would improve the model's ability to generalize to previously unseen actuators, additional descriptive features or an initial calibration procedure would likely be required to recover the lost predictive performance.}

\begin{figure*}
\centerline{\includegraphics[width=0.95\linewidth]{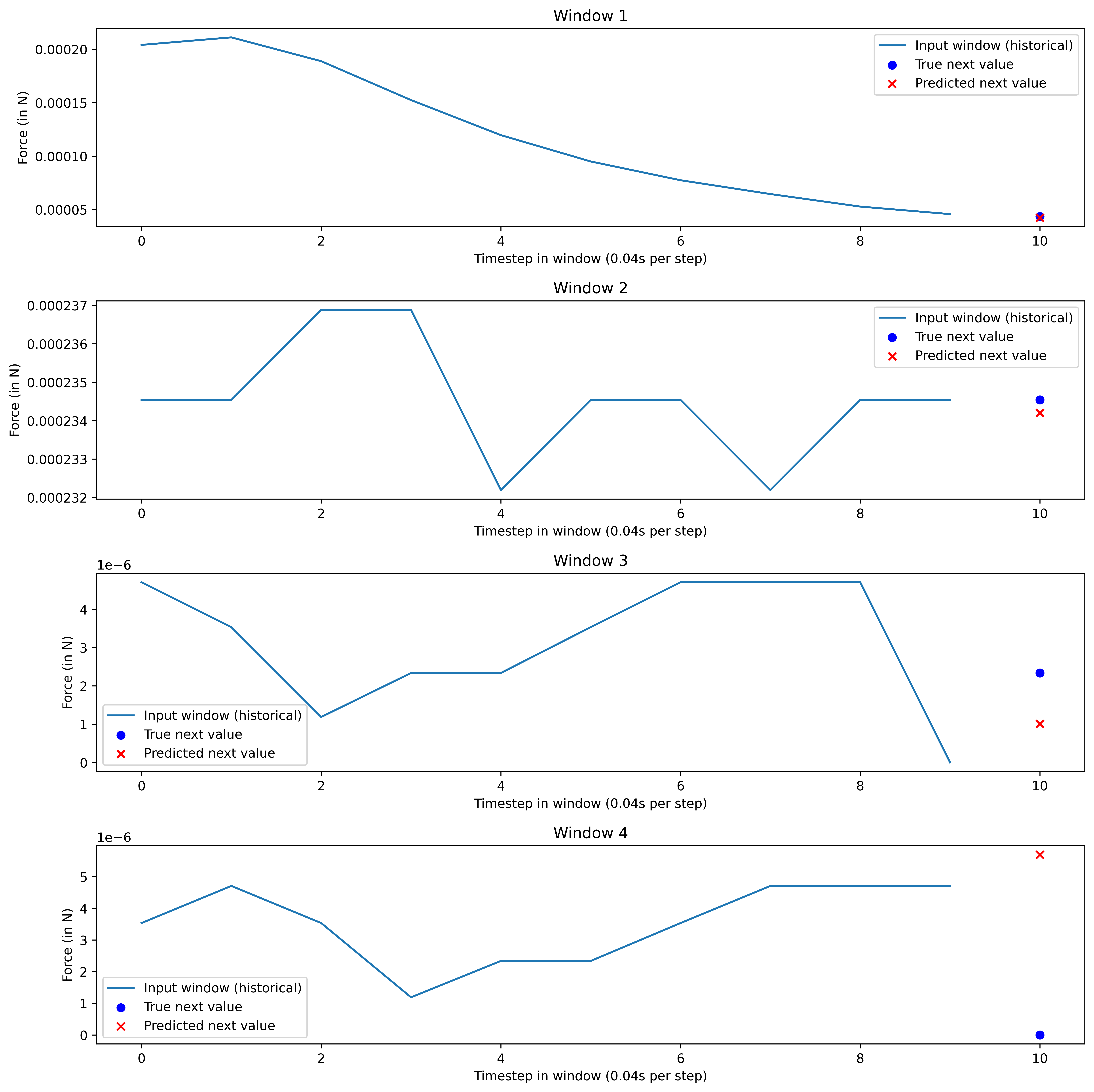}}
\caption{Four examples of input windows of 10 data points, the true next value of force and the predicted one that LSTM provides after training.\label{fig4}}
\end{figure*}

\begin{figure*}
\centerline{\includegraphics[width=0.95\linewidth]{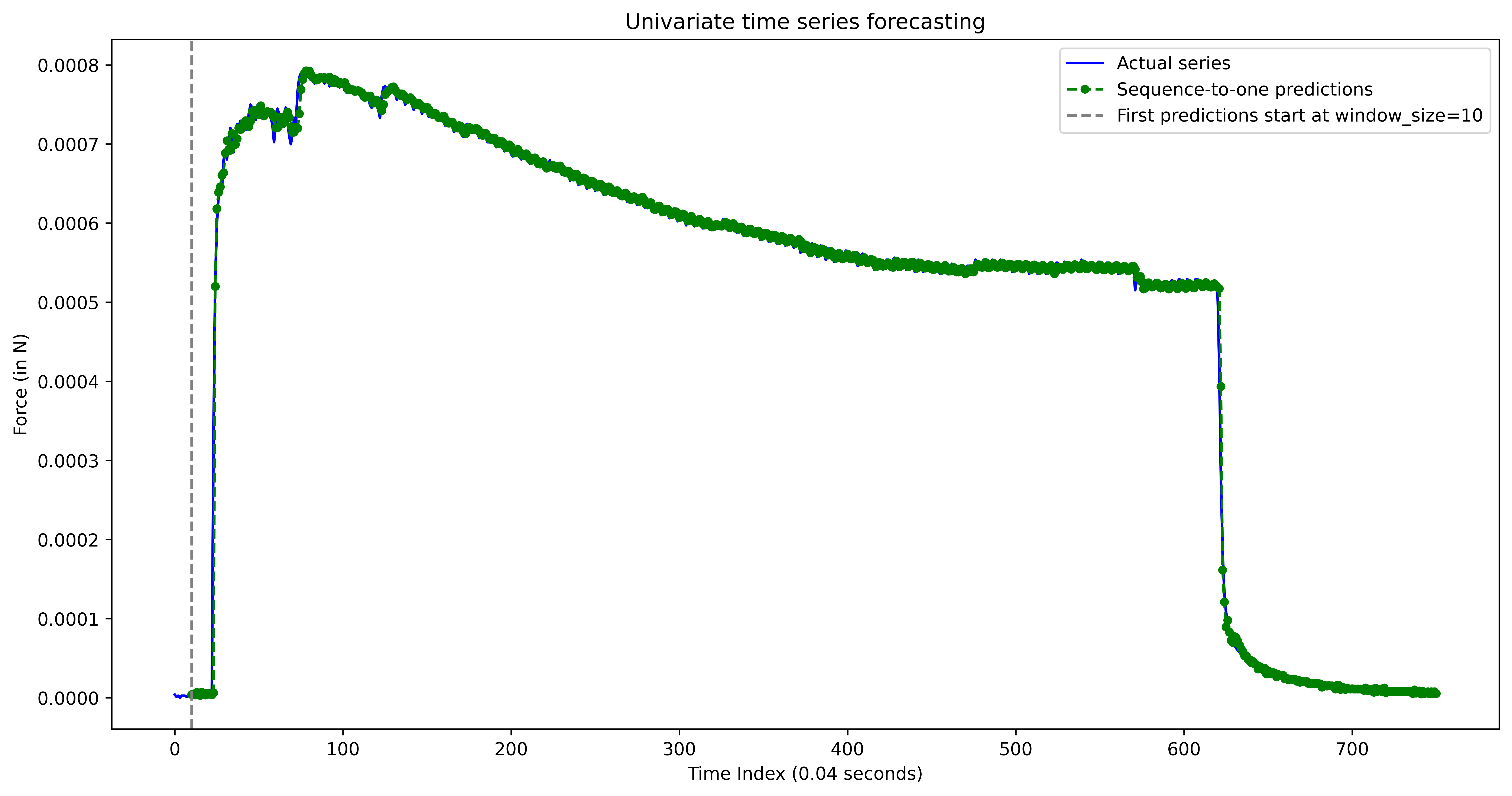}}
\caption{The actual time series for biphasic asymmetric charge‐balanced experiment with pulse modulation between $[2ms,10ms]$ at $30Hz$ and the forecast time series provided by LSTM.\label{fig5}}
\end{figure*}

\begin{figure*}
\centerline{\includegraphics[width=0.8\linewidth]{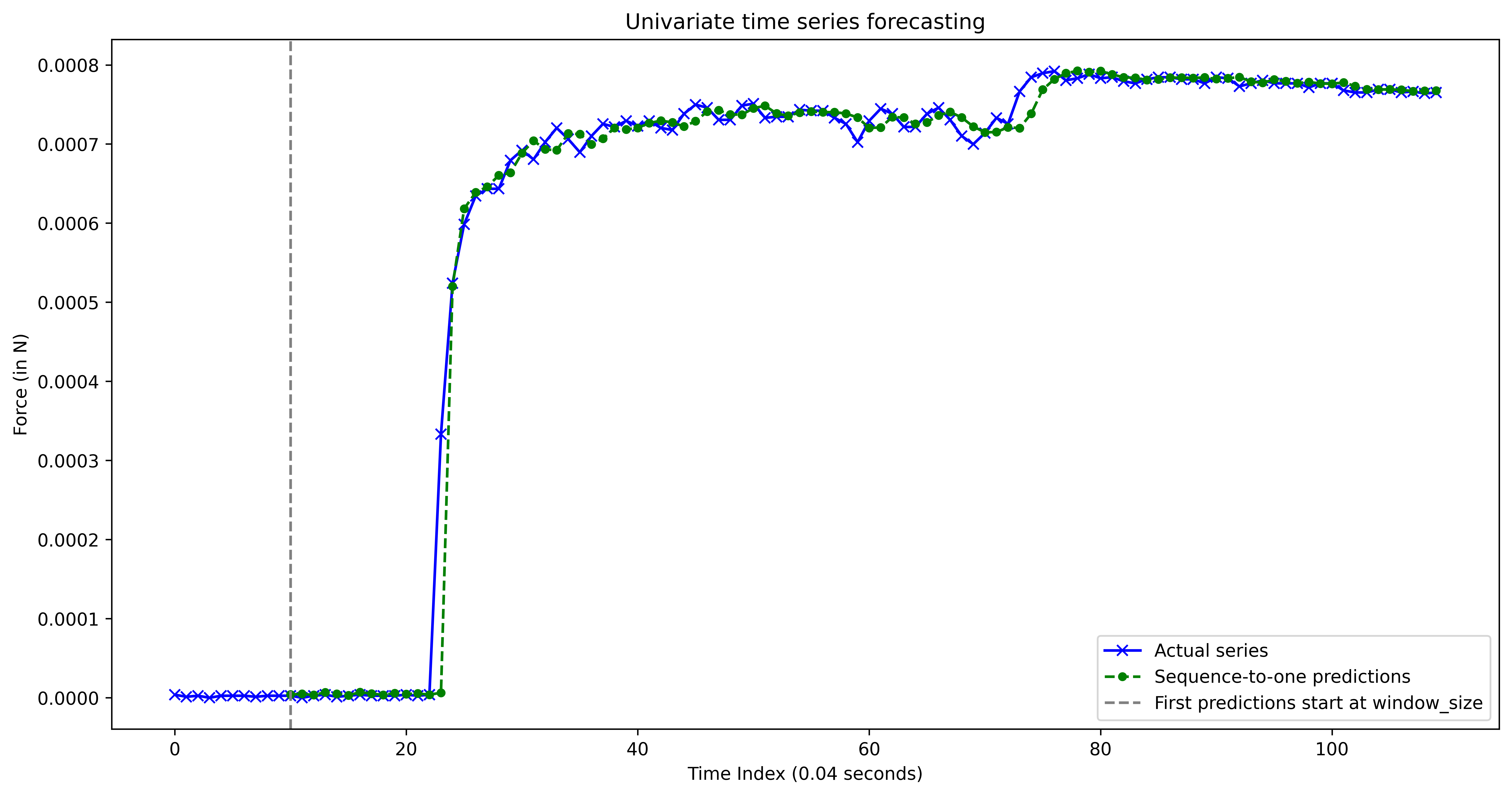}}
\caption{The first 100 time steps of the actual time series for the experiment presented in Fig \ref{fig5} and the forecast time series provided by LSTM.\label{fig6}}
\end{figure*}

For the dynamic prediction model, the dataset consisted of {a collection of data files (i.e., 17 files) per experimental method (mainly categorized by the type of stimulation signal's waveform) that totals into} 122176 training windows, each containing 10 consecutive force measurements. {The data files were split into 65\% for training (i.e. 11 files), 17.5\% for validation (i.e. 3 files) and 17.5\% for testing (i.e., 3 files). The choice to split the data set in recording files and then build the timeseries windows was motivated by the attempt to alleviate any possibility of data leakage. } After training the LSTM for 30 epochs {and for 5 different random seeds, the resulting mean MSE was 0.000458, while the mean $R^2$ score was equal to 0.9944 with standard deviation 0.002779.} Figure \ref{fig4} presents an example of four input windows, comparing the predicted next force value with the corresponding true value. Also, the complete time series of the force recorded during one specific experiment (biphasic asymmetric charge‐balanced with pulse modulation between $[2ms,10ms]$ at $30Hz$) is presented in Fig. \ref{fig5} along the forecast of the LSTM of the time series using the actual values for each sequence-to-one prediction. A close up of the initial application of the electrical stimulus for the same experiment and the equivalent LSTM forecast is provided in Fig. \ref{fig6}.

The results indicate that forecast accuracy is high and stable during steady-state operation (before stimulation application or during), but deteriorates at sudden changes. This fact suggests that the model is requiring further improvements to predict time steps when the stimulation is initially applied or changed drastically. This temporal variation highlights the importance of incorporating dynamic uncertainty into the {data-driven predictive model}.

{A sensitivity analysis was performed on the LSTM input window length by considering windows of 5, 10, and 20 samples. The resulting predictive performance remained high across these configurations (as illustrated in Table \ref{tab4}), indicating that the model is relatively robust to reasonable variations of this hyperparameter. While a comprehensive study of architectural choices and hyperparameter optimization is beyond the scope of this proof-of-concept work, these results suggest that the reported performance is not strongly dependent on the specific selection of the length of the input window.}

\begin{table*}[!h]%
\centering %
\caption{Results of the sensitivity analysis of the accuracy (in terms of $R^2$ values) of LSTM architecture with variable input window length.\label{tab4}}%
\begin{tabular}{l|lllll|l|l}
                             & \multicolumn{5}{l|}{\textbf{Random seed run}}                  &               &              \\
\textbf{Input window length} & \textbf{1} & \textbf{2} & \textbf{3} & \textbf{4} & \textbf{5} & \textbf{Mean} & \textbf{Std} \\ \hline
5                            & 0.99533951 & 0.99003184 & 0.99561816 & 0.99722385 & 0.99352944 & 0.994349      & 0.002456     \\
10                           & 0.99508995 & 0.98956478 & 0.99584907 & 0.99784845 & 0.99352038 & 0.994375      & 0.002779     \\
20                           & 0.99458599 & 0.9896276  & 0.99643332 & 0.99787599 & 0.99338758 & 0.994382      & 0.002830    
\end{tabular}
\end{table*}


\section{Conclusions}\label{sec5}

BHMs can offer unique advantages for soft robots, despite the fact that reliable characterization and control of these systems is still a major challenge. 
{While deterministic modeling approaches are in principle applicable to BHMs, their practical implementation is complicated by pronounced biological variability and complex nonlinear dynamics. In this context, ML tools can play a complementary role in supporting design, optimization, and controllability of BHMs.} The utilization of ML based approaches constitutes a first step towards better understanding their behavior and realizing more accurate digital twins. Thus, appropriate response based on varying electrical stimuli can be anticipated and real-time control methodologies can be developed.

This work demonstrated that ML models can be utilized to effectively predict BHMs outputs. We demonstrated this by training static and dynamic models on a dataset extracted from laboratory experiments with muscle rings that were stimulated electrically. The static prediction models achieved increasing accuracy in predicting the maximum force per experiment based on inputs like the sample number ID of the muscle ring, the frequency, pulse width and waveform type of stimulation signals. Whereas, the inclusion of baseline exerted force before the application of electrical stimulus enhanced the performance of a NN model by reducing the RMSE by 3.5\% relative to the initial NN configuration and by 10.6\% compared to a fine-tuned RFR model. Moreover, residual analysis confirmed random and homogeneous error distribution. The dynamic prediction model, based on a recurrent NN (i.e., LSTM architecture), reproduced the full time series of force achieved by the actuator with an $R^2$ score of 0.9944, effectively capturing the temporal behavior of the biological system.

{The distribution of stimulation parameters in the dataset is not uniform, as experimental conditions were selected based on human-expert-driven exploration and the observed actuator response \cite{collu2025development}. In particular, parameter combinations yielding consistently low force outputs were not extensively sampled. Despite this imbalance, the predictive performance of the models does not appear to be adversely affected. As observed in the predicted versus actual force plots (Figs. \ref{fig1}, \ref{fig2} and \ref{fig3}), the models maintain high accuracy across the full range of forces, and especially the lower-force regime, where data points remain closely aligned with the identity line. Nonetheless, predictions in sparsely sampled regions of the parameter space should be interpreted with caution, and further data collection would be beneficial to improve coverage and robustness.}

Nevertheless, the present study is an initial step toward implementing ML onto biohybrid systems investigation and has several limitations. Firstly, the dataset size remains relatively small for static models, since 161 experiments were performed with a singular configuration of several samples of biological actuators (i.e., muscle rings anchored around two pillars on a Petri dish with the same dimensions).
The generalization capability of the proposed models depends on the type of variability considered. With respect to actuator geometry, this remains the primary limitation of the present study. The models were trained using data from a single geometrical configuration (i.e., a muscle ring anchored around two pillars on a Petri dish), whereas it is well established that geometry and its dimensions is a key factor influencing the mechanical response of biohybrid machines. As such, the ability of the trained models to generalize to different actuator geometries is expected to be very limited.

Nonetheless, biological variability is not strongly captured in the dataset, solely due to the use of multiple biological replicates. Given that variability across samples is inherently present, despite efforts to maintain consistent laboratory techniques, the models were exposed to this variation during training. Therefore, based on the comparison of the results, the model without the sample ID may retain some robustness to biological variability; however, its ability to generalize to previously unseen biological samples has not been established. On the other hand, regarding experimental conditions, the dataset includes a range of stimulation parameters, such as frequency, pulse width, and waveform. This allows the models to learn the system response within the explored parameter space, and thus generalization is expected to be adequate within these conditions. However, extrapolation beyond the range of observed parameters remains uncertain and would require additional data.

Moreover, for dynamic models univariate time series forecasting was implemented in only an open-loop context. A closed-loop feedback implemented in the ML model would contribute a more robust methodology. Nonetheless, the use of multiple input variables (i.e., the stimulation electrical signal time series) would reinforce the predictive capacity of the model, especially in areas of sudden changes (i.e., the initial time point of stimulation application, where the current model can not tackle appropriately as realized in Fig. \ref{fig6}). These constraints limit the ability of the models to generalize and be directly deployed in more complex biohybrid systems.

As aspects of future work, the inclusion of experiments with different configurations and in larger numbers to the training dataset will be considered. Furthermore, the extension of dynamic ML models to include multiple inputs (i.e., the time series of the stimulation electrical signal as mentioned previously) will be investigated, so that models will better anticipate sudden changes in the state of the system. In addition, the inclusion of closed-loop ML methodologies will be studied to enable self-adaptive control strategies of BHMs that will be possible to directly deploy in hardware \cite{collu2025development}. The aforementioned approaches will strengthen ongoing research on BHMs and facilitate their progression beyond proof-of-concept toward full realization.

\section{Experimental Section}\label{sec3}

\subsection{3D Bioengineered Muscle Tissue}

C2C12 mouse myoblasts were purchased from ATCC and cultured in T‐175 flasks. Once cells reached 80\% of confluency, they were trypsinized and mixed with a custom-made hydrogel (30\% v/v Matrigel (Corning), 50\% v/v of 8mg/mL Fibrinogen (Sigma-Aldrich) in 1X PBS, 4\% v/v of 4U/mL thrombin (Sigma-Aldrich) in 1X PBS, and 16\% v/v of growth media (GM) supplemented with Aminocaproic Acid) at a concentration of $10 \times 10^6$ cells/mL. The composition of GM is DMEM (Gibco), 10\% FBS (Gibco), 200 nM l‐glutamine (Gibco), 1\% penicillin–streptomycin (Gibco). Then, this cell-laden hydrogel was mold casted in PDMS-based circular rings and there cultured for 2 days in growth media supplemented with 1mg/mL Aminocaproic Acid. Afterwards, the tissue was transferred to a Petri dish containing two notches in differentiation media (DM). Previous studies have showcased that utilizing circular-shaped molds for fabricating muscle actuators enables high scalability and facilitates their easy integration into various flexible structures, such as serpentine-like skeletons or posts.

The 3D circular shaped skeletal muscle tissue was maintained in the two-post system for at least 7 days during the differentiation process. DM was composed of DMEM supplemented with 10\% horse serum (Gibco), 200 $nM$ l-glutamine (Gibco), 1\% penicillin–streptomycin (Gibco), Insulin-like growth factor 1 (IGF-1, 50 $ngmL^{-1}$; Sigma-Aldrich), and 1 $mg mL^{-1}$ of ACA. Following previously published protocols, the distance between the centers of the two posts was set to 9 mm. This is a configuration that supported skeletal muscle cell differentiation and allowed performance evaluation of tissues in terms of contraction force, as the bending of the pillars can be analyzed to calculate the force exerted by the muscle against the pillars during contraction \cite{collu2025development}.

\subsection{Electrical Stimulation}

The muscle tissues, assembled in the same two‐post systems used during differentiation, were electrically stimulated with a train of pulses. The handmade electrodes are two cylindrical graphite rods (cat. number 30250, Ladd Research) placed 13 mm apart on opposite sides of the Petri dish cover (Fig. \ref{fig_ric}C). The portable stimulator consists of two stacked custom PCBs and a commercial microcontroller unit. Each PCB is a two‐layer board, identical in size and profile, designed to mount onto the ESP32‐WROOM‐32UE development MCU, a dual‐core 32‐bit microprocessor with Bluetooth and USB functionality (Fig. \ref{fig_ric}B).

The device is capable of providing electrical stimulation in current-control mode (CCM) with four fully programmable stimulation waveforms based on monophasic and biphasic shapes (monophasic, symmetric biphasic, charge‐balanced asymmetric biphasic, and symmetric triangular - Fig. \ref{fig_ric}A). Waveforms’ shapes can be modified in terms of frequency,  pulse-width duration (PW), and amplitude through a custom Matlab Graphical User Interface (GUI). Electrical stimulation algorithms based on amplitude, pulse width, or frequency modulation can be programmed through the GUI. Electrical stimulation modulation can also be implemented using linear steps or a trapezoidal pattern, enabling investigation of advanced control strategies useful for adapting stimulation to the different biological tissues used as actuators. 

{The stimulation frequencies reported in Table \ref{stim_summary} were selected based on the fact that they lie within the range commonly used in the literature \cite{hamaguchi2022establishment,nasledov2007changes,tamura2020electrically}, where stimulation frequencies are typically at or below 100 Hz, while being sufficiently high to induce and maintain tetanic muscle contraction. During tetanic stimulation, the stimulus frequency must be high enough to prevent relaxation between successive contractions, resulting in a sustained muscle force. The contractile dynamics are considerably slower than the electrical stimulation, producing a predominantly low-frequency and quasi-continuous force response throughout the stimulation period.}

\subsection{Biohybrid Actuator Stimulation Procedures}

Three different stimulation strategies were applied to verify the ability to induce tetanic contractions and to modulate actuator contraction in a controllable manner.

Twitch contraction was tested on two different muscle actuators, while stimulation was provided at 1 Hz for 10 s.

The first bioactuator was stimulated with the four available waveforms in the stimulator, using a pulse width (PW) of 5 $ms$ and an amplitude of 18 $mA$. On the second biohybrid actuator, the effect of injected stimulation charge on contraction strength was studied by fixing the amplitude but using different PWs (2 $ms$, 20 $ms$, 50 $ms$).

Tetanic contraction was tested by stimulating the biohybrid actuators for 10 s at three frequencies: 20, 30, and 50 Hz, while maintaining an amplitude of 18 mA. Stimulation was performed using the four available waveforms in the stimulator (biphasic symmetric, biphasic asymmetric charge‐balanced, monophasic, and triangular biphasic charge‐balanced), with the PW fixed according to the frequency.

The dynamic modulation of contraction was investigated using staircase or linear modulation of stimulation parameters.

The modulation was implemented separately for frequency, amplitude, and PW.

\begin{figure*}
\centerline{\includegraphics[width=\linewidth]{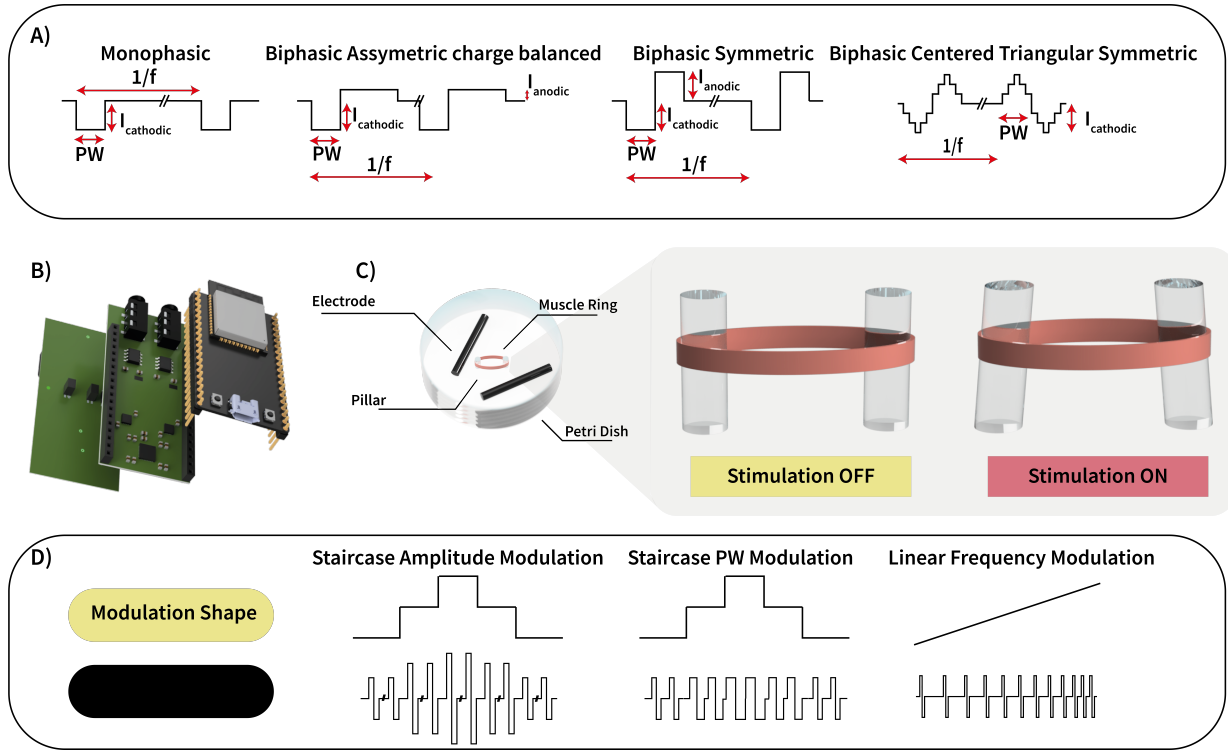}}
\caption{Electrical stimulation patterns and setup. (A) Stimulation was performed using four different waveforms. (B) Stimulation was performed using a portable stimulator \cite{collu2025development}. (C) The biohybrid actuator was placed inside a Petri dish and mounted on two movable pillars. Electrical stimulation caused the muscle bioactuator to contract, inducing movement of the pillars. (D) The stimulation parameters were modulated using linear modulation following a ramp or staircase pattern.\label{fig_ric}}
\end{figure*}

Staircase modulation was used separately for amplitude and PW. During this variation mode, the modulated parameter (i.e., amplitude or PW) is modified linearly between a minimum value and a maximum value through a defined number of steps, as shown in Fig. \ref{fig_ric}D. The rate of increase due to modulation depends on the minimum and maximum values of the modulated parameter and the number of steps selected by the user.

Linear ramp modulation was used separately for frequency, amplitude, and PW. During this modulation, the modulated parameter (i.e., Amplitude, PW, or Frequency) is increased linearly after each stimulation pulse until the maximum parameter is reached. The rate of increase due to modulation depends on the parameter's minimum and maximum values and the time interval over which it varies.

A summary of the stimulation conditions is presented in Table \ref{stim_summary}.

\begin{table*}
\centering
\caption{Summary of experiments\label{stim_summary}}%
\resizebox{\textwidth}{!}{%
\begin{tabular}{|c|c|c|c|c|c|c|c|c|}
\hline
Aim & \begin{tabular}[c]{@{}c@{}}IDs of  \\ Actuators\end{tabular} & Modulation & Freq {[}{Hz}{]} & PW {[}ms{]} & Waveform & Amplitude {[}mA{]} & Stim. Time & N. Measures \\ \hline

\begin{tabular}[c]{@{}c@{}}\\    \\ Tetanic\\  contraction  \\ \end{tabular}    & \begin{tabular}[c]{@{}c@{}}S4\\    \\ S5\\    \\ S6\end{tabular} & {No Modulation} & \begin{tabular}[c]{@{}c@{}}20\\    \\ 30\\    \\ 50\\    \\ 70\end{tabular} & \begin{tabular}[c]{@{}c@{}}MAX\\    \\ PW\end{tabular} & \begin{tabular}[c]{@{}c@{}}Mono\\    \\ Triangular\\    \\ Asym.\end{tabular} & 18 & 10 s & 31 \\ \hline

\begin{tabular}[c]{@{}c@{}}\\  Contraction  \\ Modulation\\    \\ \end{tabular}    & \begin{tabular}[c]{@{}c@{}}S1\\    \\ S2\\    \\ S3\end{tabular} & \begin{tabular}[c]{@{}c@{}}\\  Linear  \\ Amplitude\\  Modulation  \\ \end{tabular}      & 50 & 5 & \begin{tabular}[c]{@{}c@{}}Sym.\\    \\ Asym\end{tabular} & 0-18 & 80 s & 5 \\ \hline

\begin{tabular}[c]{@{}c@{}}\\ Contraction\\ Modulation\\    \\  \end{tabular}    & \begin{tabular}[c]{@{}c@{}}S7\\    \\ S8\\    \\ S9\end{tabular} & \begin{tabular}[c]{@{}c@{}}Staircase\\  Amplitude  \\    Modulation\end{tabular} & \begin{tabular}[c]{@{}c@{}}20\\    \\ 30\\    \\ 50\end{tabular} & MAX PW & \begin{tabular}[c]{@{}c@{}}Mono\\    \\ Sym\\    \\ Triangular\\    \\ Asym.\end{tabular} & 3 - 18 & \begin{tabular}[c]{@{}c@{}}6 steps \\    \\ 2 s\end{tabular} & 39 \\ \hline

\begin{tabular}[c]{@{}c@{}}\\ Contraction\\ Modulation\\    \\  \end{tabular} & \begin{tabular}[c]{@{}c@{}}S13\\    \\ S14\\    \\ S15\end{tabular} & \begin{tabular}[c]{@{}c@{}}Staircase PW\\    \\ Modulation\end{tabular} & \begin{tabular}[c]{@{}c@{}}20\\    \\ 30\\    \\ 50\end{tabular} & 2ms – MAX & \begin{tabular}[c]{@{}c@{}}Mono\\    \\ Sym.\\    \\ Triangular\\    \\ Asym.\end{tabular} & 18 &  & 39 \\ \hline

\begin{tabular}[c]{@{}c@{}}\\ Contraction\\ Modulation\\    \\  \end{tabular}& \begin{tabular}[c]{@{}c@{}}S13\\    \\ S14\end{tabular} & \begin{tabular}[c]{@{}c@{}}Linear PW\\    \\ Modulation\end{tabular} & \begin{tabular}[c]{@{}c@{}}30\\    \\ 35\\    \\ 50\end{tabular} & \begin{tabular}[c]{@{}c@{}}0.1ms -MAX\\    \\ 2-MAX\end{tabular} & Sym. & 18 & \begin{tabular}[c]{@{}c@{}}10 s\\    \\ 20 s\end{tabular} & 7 \\ \hline

\begin{tabular}[c]{@{}c@{}}\\ Contraction\\ Modulation\\    \\  \end{tabular} & \begin{tabular}[c]{@{}c@{}}S10\\    \\ S11\\    \\ S12\end{tabular} & \begin{tabular}[c]{@{}c@{}}Linear\\    \\ Frequency\\    \\ Modulation\end{tabular} & \begin{tabular}[c]{@{}c@{}}10-20\\    \\ 10-30\\    \\ 10-50\\    \\ 20-25   \\ \\ 20-35\\    \\ 35-50\end{tabular} & Max PW & \begin{tabular}[c]{@{}c@{}}Mono\\    \\ Sym\\    \\ Triangular\\    \\ Asym.\end{tabular} & 18 & 20 s & 40 \\ \hline
\end{tabular}%
}
\end{table*}


\subsection{ML-based Methods}

 The stimulation procedures mentioned in the previous Section produced a total of 161 separate experiments. The overall dataset consists of 123786 time steps where force is measured. The biohybrid actuator was recorded using the Leica Thunder imaging system to determine the bending of the two posts under specific electrical stimulation \cite{collu2025development}. The measurement interval throughout all experiments was $0.04s$. {To ensure reproducibility, the full implementation of models\footnote{\url{https://github.com/Antisthenis/biomeld_actuator_dnn}} and the dataset\footnote{\url{https://zenodo.org/records/22895110}} used for this study are publicly available.
 The repository includes detailed information on model architectures, training procedures, hyperparameters (e.g., learning rates), weight initialization strategies, and the software frameworks used.}

\subsubsection{Static Prediction Models}
\label{Static_methods}

{For static prediction models the dataset has four inputs (two numerical and two categorical) and one output - target. The numerical inputs are considered the frequency and pulse width of the stimulation signals. The categorical inputs are the sample number (one of the 15 identical muscle tissue rings mold casted) and the waveform type of the stimulation signal applied to the muscle rings per experiment (all possible combinations between the ``Waveform'' and ``Modulation'' columns of Table \ref{stim_summary}). The categorical inputs are discrete labels without an inherent mathematical scale, thus they are encoded using one-hot encoder method. The output is the maximum achieved force exerted on the pillars throughout the duration of an experiment.}

As an initial approach a RFR was implemented due to its robustness, simplicity, and proven effectiveness in handling complex datasets. A RFR is an ensemble learning method \cite{breiman2001random} that constructs multiple decision tree regressors using different bootstrap samples of the training data and combines their predictions through averaging. This ensemble approach reduces variance compared to a single decision tree, improving predictive accuracy and minimizing the risk of overfitting.

To optimize the performance of the RFR, hyperparameter tuning was conducted using randomized search cross validation. This approach samples a fixed number of parameter settings from specified distributions and evaluates model performance through cross validation, providing an efficient alternative to exhaustive grid search. The hyperparameters explored included: (i) the number of estimators (i.e., number of trees in the forest) within the set \{100, 200, 300\}, (ii) maximum depth of individual trees within \{4, 6, 8, 10\}, (iii) minimum samples split, defining the minimum number of samples required to split an internal node, within \{2, 5, 10\}, (iv) and minimum samples leaf, indicating the minimum number of samples required to be at a leaf node, within \{1, 2, 4\}. The final model parameters were determined through a cross validated search over these combinations, ensuring that the selected configuration yielded the optimal balance between bias and variance.

As a second modeling approach, a feedforward NN was implemented using the same input and output variables as previously. The NN was selected because it can flexibly approximate complex, nonlinear relationships and automatically learn interactions among features in a data-driven fashion, which may be difficult to capture with tree-based methods. In particular, NNs may excel when the underlying mapping from inputs to outputs is highly nonlinear, continuous, or involves subtle cross-feature dependencies that ensembles of piecewise constant trees might not capture well.

All hidden layers employed the ReLU (rectified linear unit) activation function. The network architecture consisted of four hidden layers with 68, 50, 30, and 10 neurons respectively, followed by a single output neuron to predict the maximum force. Before training, numerical features were standardized and categorical features were transformed via one-hot encoding, as performed for the RFR model. The Adam optimizer was used for training, which proceeded for 50 epochs.

In addition, as an optimized approach the NN implementation was improved by utilizing the baseline force observed in the muscle rings during the first 0.4 seconds of each experiment as an auxiliary input feature. This step was motivated by the observed considerable variation in baseline force across experiments and muscle samples; by supplying this baseline explicitly, the network could better adjust for inter-sample offsets. All other hyperparameters and training settings remained consistent with those described above.

\subsubsection{Dynamic Prediction Models}

For the dynamic prediction models, the objective was to emulate the behavior of the BHM model so as to reproduce its outputs in a time series fashion. Thus, LSTM networks were selected. LSTMs are a specialized recurrent NN architecture that can capture long term dependencies in sequential data by using gating mechanisms (input, forget, and output gates) to regulate information flow, thereby mitigating the vanishing (and exploding) gradient problems encountered in standard recurrent NN. This makes LSTMs well suited for modeling temporal dynamics and producing stable time series predictions over multiple time steps.

The dataset was segmented into overlapping sequences, i.e., windows, of 5, 10 and 20 time steps (each with a duration of 0.04 seconds), with a batch size of 32, resulting in 122176 training windows. Prior to training, all features were linearly scaled, mapping the minimum and maximum observed values to a fixed range without disproportionately attenuating the influence of outliers. The time series data were then partitioned into training, validation, and testing sets in proportions of 65\%, 17.5\%, and 17.5\%, respectively. The LSTM architecture comprised a single recurrent layer of 64 hidden units, followed by a fully connected layer of 32 neurons using ReLU activation, which in turn fed into a final output neuron. The network was trained for 30 epochs.



\subsubsection*{Acknowledgments}
This project has received funding from the European Union’s Horizon Europe research and innovation programme under grant agreement No. 101070328. M.-A.T., M.P.H. and A.A. were funded by the UK Research and Innovation grant No. 10044516.
\\
The authors acknowledge the use of ChatGPT (OpenAI; version: GPT-5.6) throughout the preparation of this manuscript for spelling, grammar checking and minor language editing. The tool was used during manuscript preparation up to September 2026. AI-assisted language corrections were applied throughout the manuscript, without targeting any specific section. All AI-generated suggestions were manually reviewed and validated by the authors for accuracy and consistency with the original content. No sensitive, proprietary, or human-subject data were provided to the AI tool.


\subsubsection*{Conflict of interest}

The authors declare no potential conflict of interests.

\appendix

\section{Short text and graphic for the Table of Contents\label{app1}}
Skeletal muscle-based biohybrid actuators anchored on flexible polymer pillars were studied under static and dynamic machine learning prediction models. Both modeling approaches demonstrate high predictive accuracy with the best performance of the static models, i.e. Neural Network regressor characterized by $R^2$ of 0.9346, whereas the dynamic model, i.e. Long Short-Term Memory network achieved $R^2$ of 0.9944.

\begin{figure*}[!h]
\centerline{\includegraphics[width=\linewidth]{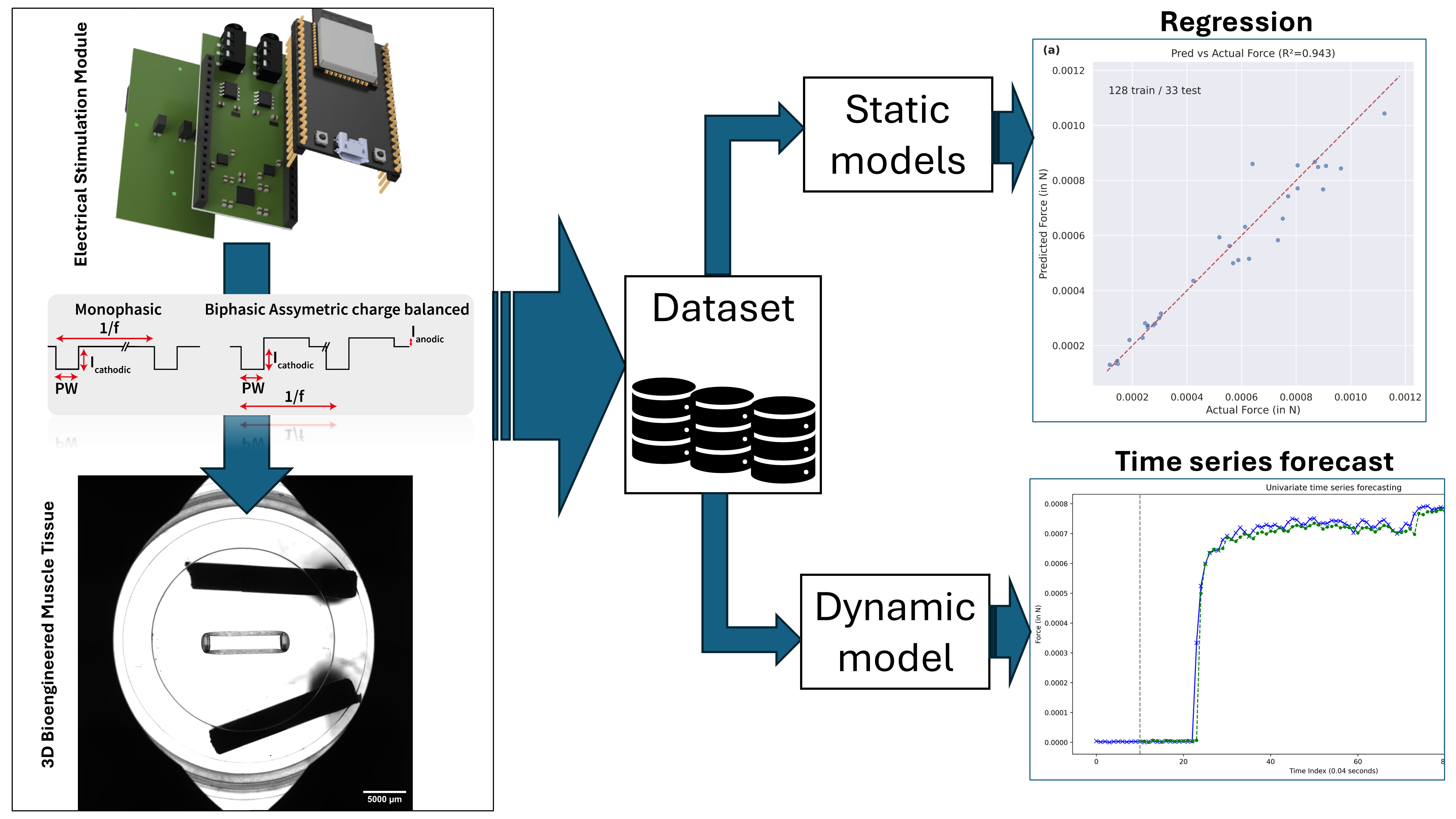}}
\end{figure*}

\end{document}